\documentclass[letterpaper]{article} % DO NOT CHANGE THIS
\usepackage[accepted]{icml2019}  % DO NOT CHANGE THIS
\usepackage{times}  % DO NOT CHANGE THIS
\usepackage{helvet} % DO NOT CHANGE THIS
\usepackage{courier}  % DO NOT CHANGE THIS
\usepackage[hyphens]{url}  % DO NOT CHANGE THIS
\usepackage{graphicx} % DO NOT CHANGE THIS

\usepackage{microtype}
\usepackage{graphicx}
\usepackage{subfigure}
\usepackage{multirow}
\usepackage{booktabs} % for professional tables

\usepackage{hyperref}
\usepackage{amsmath}

\DeclareMathOperator*{\argmax}{arg\,max}

\usepackage{algorithm}
\usepackage{algorithmic}
\usepackage{graphicx}  % DO NOT CHANGE THIS
\title{Context-aware Active Multi-Step Reinforcement Learning }
\author{\Large \textbf{Gang Chen\textsuperscript{\rm 1}, Dingcheng Li\textsuperscript{\rm 2}, and Ran Xu\textsuperscript{\rm 3}}\\ % All authors must be in the same font size and format. Use \Large and \textbf to achieve this result when breaking a line
\textsuperscript{\rm 1}Department of Computer Science, SUNY at Buffalo\\ %If you have multiple authors and multiple affiliations
\textsuperscript{\rm 2}Baidu\\
\textsuperscript{\rm 3}Salesforce\\
gangchen@buffalo.edu \quad dingchengl@gmail.com \quad xurantju@gmail.com  % email address must be in roman text type, not monospace or sans serif
}
 
\begin{document}

\maketitle

\begin{abstract}
Reinforcement learning has attracted great attention recently, especially policy gradient algorithms, which have been demonstrated on challenging decision making and control tasks. %REINFORCE algorithm is easy and effective, but it has high variance and requires large number of episodes to make it stable. Actor-critic algorithm can be trained with much less data, but has very high computation cost on each time step, which is explicitly significant while using deep neural network as function approaximator. 
In this paper, we propose an active multi-step TD algorithm with adaptive stepsizes to learn actor and critic. Specifically, our model consists of two components: active stepsize learning and adaptive multi-step TD algorithm. Firstly, we divide the time horizon into chunks and actively select state and action inside each chunk. Then given the selected samples, we propose the adaptive multi-step TD, which generalizes TD($\lambda$), but adaptively switch on/off the backups from future returns of different steps. Particularly, the adaptive multi-step TD introduces a context-aware mechanism, here a binary classifier, which decides whether or not to turn on its future backups based on the context changes. Thus, our model is kind of combination of active learning and multi-step TD algorithm, which has the capacity for learning off-policy without the need of importance sampling. We evaluate our approach on both discrete and continuous space tasks in an off-policy setting respectively, and demonstrate competitive results compared to other reinforcement learning baselines. 
\end{abstract}

\section{Introduction}
\label{submission}
REINFORCE\cite{Williams92}, as the most classic policy gradient approach \cite{SuttonB98,Kakade2001,Schulman2015}, directly optimizes the parameterized policy to improve the total return. The advantages of policy gradient algorithm is that it models the policy effectively by optimizing the parameters related to the policy control, and it handles high dimensional and continuous actions well. However, it suffers from high variance by delaying its model update and leading to slow learning until the end of the episode. Value-based (or critic only) approaches, such as Q-learning \cite{Watkins92}, iteratively improve its evaluations of the action-state pairs, which in turn can guide an agent what action to take under what circumstances. The updating procedure derived from Bellman equation needs to handle every state and action from environment, which is computationally intensive especially for the continuous action space. Actor-critic algorithms \cite{Witten77,Konda00} leverage both advantages of policy gradient and value function approaches\cite{SuttonB98,Kakade2001,Schulman2015,SchulmanWDRK17}, such as lower variance and good convergence properties. Thus, actor-critic algorithms have gained popularity in reinforcement learning community recently. For example, Lillicrap et al. extends the actor-critic model using deep learning to handle the continuous action space \cite{LillicrapHPHETS15}. Asynchronous Advantage Actor Critic ($A^3C$) \cite{Mnih2016} presents a asynchronous version of actor critic algorithm, which uses parallel actor learners to update a shared model to make the learning process stabilized. Schulman et al. proposed a generalized advantage estimation (GAE) uses an exponentially-weighted estimator \cite{Schulmanetal2016} of the advantage function to reduce the variance of policy gradient. Recently, Twin Delayed Deep Deterministic Policy Gradients (TD3) \cite{FujimotoHM18} uses the minimum of two critics to limit the overestimated bias in actor-critic network. A soft actor critic algorithm \cite{HaarnojaZAL18}, an off policy actor-critic, leverages the maximum entropy to balance the exploration and exploitation in reinforcement learning framework. However, actor-critic as a special TD algorithm with one step \cite{Sutton1988} updates the model (actor and critic) with every time step \cite{LillicrapHPHETS15,HaarnojaZAL18}, which is time consuming and computationally intensive while updating a large deep model controller. %Recently, Nachum et al. prove the equality between Value and Policy Based Reinforcement Learning \cite{NachumNXS17}. 
Multi-step methods take a balance strategy between actor-critic ( or TD(0)) and Monte Carlo, and have gained resurgence in reinforcement learning \cite{AsisHHS18,Hernandez19}, such as n-step Tree Backup \cite{PrecupSS00}, Retrace($\lambda$) \cite{MunosSHB16}, n-step Q($\sigma$) \cite{AsisHHS18} and Sarsa \cite{Rummery94}. Unfortunately, the step-size must be manually set up as the TD($\lambda$) \cite{Sutton1988}.

Instead of updating the model at each state, in this paper, we propose to actively learn the time step in the outer loop while internalizing the multi-step TD algorithm in the inner loop. %The purpose of active time step optimization is to find the significant state in an given interval and then use the multi-step TD algorithm to update the new policy. 
However, it is a challenge to optimize the step size, considering the time step is a discrete variable. To do it, we divide the trajectories into intervals. Then, our method can select the most significant states and actions inside of each interval, which implicitly learns the step size in the outer loop. While in the inner loop, we propose the adaptive multi-step TD learning, which turns on/off the future backups if its context changes dramatically. In particular, we introudce a binary classifier for the future backups. Based on this function, our model generalizes TD($\lambda$) to adaptively detect the context change and take average of backups of different steps in the lookahead environment, and then update the model with the average target value to improve the training effectiveness. 

%In this paper, we propose to actively learn the time step in the outer loop while internalizing the actor-critic algorithm in the inner loop. 
%related to active learning
%related to meta-learning 
In the experiments, we demonstrate that our active multi-step TD algorithm does in fact perform well by a wide margin on a bunch of continuous control tasks, compared to prior reinforcement learning methods.

\section{Problem Setting and Notation}
%\subsection{Background}
We consider the usual reinforcement learning problem (i.e. optimal policy existed) with sequential interactions between an agent and its environment \cite{SuttonB98} in order to maximize a cumulative return. At every time step $t$, the agent selects an action $a_t$ in the state $s_t$ according its policy and receives a scalar reward $r_t(s_t, a_t)$, and then transit to the next state $s_{t+1}$. The problem is modeled as Markov decision process (MDP) with tuple: $(\mathcal{S}, \mathcal{A}, p(s_{0}), p(s_{t+1}|s_t, a_t),  r(s_t, a_t), \gamma )$. Here, $\mathcal{S}$ and $\mathcal{A}$ indicate the state and action space respectively, $p(s_{0})$ is the initial state distribution. $p(s_{t+1}|s_t, a_t)$ is the state transition to $s_{t+1}$ given the current state $s_t$ and action $a_t$, $r(s_t, a_t)$ is reward from the environment after the agent taking action $a_t$ in state $s_t$ and $\gamma$ is the return discount factor, which is necessary to decay the future rewards ensuring bounded returns. We model the agent's behavior with 
$\pi_{\theta} (a|s)$, which is a parametric distribution from a neural network. %The return at time $t$ is defined as the discounted sum of rewards $R_t^{T} = r_{t} + \gamma r_{t+1} + ... + \gamma^K r_{T}$.

Suppose we have the finite trajectory length while the agent interacting with the environment. The return under the policy $\pi$ for a trajectory $\tau = {  ( s_t,  a_t   )  }_{t=0}^T$   
\begin{align}\label{eq:obj}
J(\theta) & = E_{\tau \sim \pi_{\theta}(\tau)} [  r(\tau)  ] = E_{\tau \sim \pi_{\theta}(\tau)} [  R_0^T ] \nonumber \\ 
 & = E_{\tau \sim \pi_{\theta}(\tau)} [  \sum_{t=0}^T  \gamma^t r(  s_t, a_t   )   ] 
\end{align}
where $\pi_{\theta}(\tau)$ denotes the distribution of trajectories, 
\begin{align}
 \rho(\tau) &= \pi(s_0, a_0, s_1, ..., s_T, a_T) \nonumber \\
 & = p(s_0) \prod_{t=0}^T \pi_{\theta}(a_t | s_t) p(s_{t+1} | s_t, a_t) 
\end{align}
For convenience, we can absorb $\gamma$ into reward in Eq. \ref{eq:obj}. The goal of reinforcement learning is to learn a policy $\pi$ which can maximize the expected returns.
\begin{equation} \label{eq:eq1}
\theta = \argmax J(\theta) = \argmax  E_{\tau \sim \pi_{\theta}(\tau)} [  R_0^T  ]
\end{equation} 

\subsection{Policy gradient}
Take the derivative w.r.t. $\theta$
%\begin{align}
%J(\theta) = E_{\tau \sim \pi_{\theta}(\tau)} [  r(\tau)  ] \\
%= \int \rho(\tau)   r(\tau) d\tau  
%\end{align}
%where 
%\begin{align} 
%   &  \pi_{\theta}(\tau)  =      \pi_{\theta}(s_1, a_1, ..., s_T, a_T)   \\
%\Rightarrow & \textrm{log} \pi_{\theta}(\tau) = \textrm{log} p(s_1) + \sum_{t=1}^T    \textrm{log} \pi_{\theta}(a_t | s_t) +  \textrm{ log} p( s_{t+1} | s_t, a_t  )
%\end{align}
\begin{align}\label{eq:eq2}
 \nabla_{\theta} J (\theta) & =\nabla_{\theta}   E_{\tau \sim \pi_{\theta}(\tau)} [      R_0^T      ] = \nabla_{\theta}  \int \rho(\tau)    R_0^T  d\tau  \nonumber \\
&=  \int   \nabla_{\theta} \rho(\tau)    R_0^T  d\tau =  \int  \rho(\tau)  \frac{ \nabla_{\theta}  \rho(\tau) }{ \rho(\tau)  }     R_0^T  d\tau \nonumber \\
&=  E_{\tau \sim \pi_{\theta}(\tau)} [  \nabla_{\theta} \textrm{log}  \pi_{\theta}(\tau)        R_0^T      ] 
\end{align}
As we can see the high variance Eq. \ref{eq:eq2} is attributed to the stochastic rewards from trajectory. To reduce the variance, the baseline is introduced to Eq. \ref{eq:eq2}. Then we have the following policy gradient:
\begin{align}
\nabla_{\theta} J (\theta) = E_{\tau \sim \pi_{\theta}(\tau)} [  \nabla_{\theta} \textrm{log}  \pi_{\theta}(\tau)    ( R_0^T  -b)    ] 
\end{align}

If we replace the bias with value function $V(s)$, and factorize the above equation into each time step TD(0), we get
\begin{align}\label{eq:grad}
&  \nabla_{\theta} J (\theta) \nonumber \\%= E_{\tau \sim \pi_{\theta}(\tau)} [  \nabla_{\theta} log  \pi_{\theta}(\tau)       (r(\tau)  -b)    ]  \\
=& E_{\tau \sim \pi_{\theta}(\tau)}    \bigg[  \sum_{t=0}^{T}   \nabla_{\theta}   \textrm{log}  \pi_{\theta}( a_t | s_t ) \big(  Q(s_t, a_t) -  V(s_{t})  \big)  \bigg] \nonumber \\
=& E_{\tau \sim \pi_{\theta}(\tau)}    \bigg[  \sum_{t=0}^{T}   \nabla_{\theta}   \textrm{log}  \pi_{\theta}( a_t | s_t ) \big(  r(s_t, a_t) +V(s_{t+1})  -  V(s_{t})  \big)  \bigg]
\end{align}
where $Q(s_t, a_t) = \sum_{t^\prime=t}^T  E_{ \pi_{\theta} } ( r(s_t, a_t) | s_{t^\prime}, a_{t^\prime} )$, and $V(s_{t+1}) = \sum_{ {a_t}\sim  \pi_{\theta}(\tau)} Q(s_t, a_t) $. In every time step $t$, the actor-critic algorithm updates parameters for the value function $V(s)$ with temporal difference and policy gradient in Eq. \ref{eq:grad}. This gradient has lower variance, i.e. single state transition, but is biased because of the potential inaccuracy of the lookahead estimate of $V(s)$.
%further, we have in the empirical objective function
%\begin{align}
%\nabla_{\theta} J (\theta)    \approx \frac{1}{N}\sum_{i=1}^N (  \sum_{t=1}^T \nabla_{\theta} log \pi_{\theta} (   a_{i,t} | s_{i,t}  )     ) ( \sum_{t=1}^T r( s_{i, t}, a_{i, t})   )
%\end{align}
Thus, we can see that the bias-variance tradeoff is significantly influenced by the policy $\textrm{log}  \pi_{\theta}  (\tau) $ and the temporal difference $\delta_t(s_t)$
\begin{align} \label{eq:deltat}
\delta_t = \big(  r_t  +  V(s_{t+1}) \big)  - V(s_t)
\end{align}
Furthermore, given a specific time $t$, the contributions from $\pi_{\theta}( a_t | s_t )$ and $\delta_t$ varies much to the policy gradient. For example, at some state $s_t$, if the temporal difference $\delta_t$ is too small, then we can skip this step to update actor-critic.

Instead of updating at each step $t$, we split the horizon into chunks, and find the most significant state (or $t$) inside each chunk, and then we query the oracle, such as actor-critic. In turn, we can update the model inside the actor-critic. %In the following section, we introduce the coarse to fine time chunking and then we introduce our active actor critic algorithm, which can adaptively learn the stepsize. 

\subsection{Value estimation}
TD($\lambda$) is a popular TD learning algorithm \cite{SuttonB98} to estimate state-value function $V(s_{t})$, which perfectly combines one-step TD prediction with Monte Carlo methods through the use of eligibility traces and the trace-decay parameter $\lambda$. Given the current policy $\pi_{\theta}$, the true value function $V(s_t)$ at step $t$, which looks ahead to the end of episode with discounted rewards as
\begin{align}\label{eq:tdlambda}
R_t^{T} = r_{t} + \gamma r_{t+1} + ... + \gamma^n r_{t+n} + ...
\end{align}
where $r_{t}$ is the reward at $s_{t}$ and $a_{t}$, which are omitted for simplicity. For $n$ step TD($n$), we can use $n$-step expected Sarsa to handle both on-policy and off-policy by using the return
\begin{align}\label{eq:esarsa}
R_t^{n} = r_{t} + \gamma r_{t+1} + ... + \gamma^n E_{a \sim \pi(a)}(Q(s_{t+n}, a_{t+n}))
\end{align}
Note that the first $n-1$ rewards are returns of states and actions sampled according to the behaviour policy, but the last state is backed up according to the expected action-value under the target policy. Similarily as TD($\lambda$), we can average $n$-step TD for different $n$ to balance the bias and variance. Suppose we get stepsize $n_1, n_2, ..., n_l$, and look forward different timesteps ahead, then we can take the average and yield the return at time step $t$ as following
\begin{align}\label{eq:tdaverage}
R_t^{avg} = \frac{ \lambda_1 R_t^{n_{1}} + \lambda_2 R_t^{n_{2}} + ...  + \lambda_l R_t^{n_{l}} } {   \lambda_1 + \lambda_2 + ... + \lambda_l  }
\end{align}
where $\lambda_1,\lambda_2,...,\lambda_l$ are the weights repectively for TD($n_1$), TD($n_2$),..., TD($n_l$).

One issue arising for multi-step reinforement learning is that we can't assume that these trajectories would have been taken if the agent was using the current policy. In other words, we may need off-policy correction or importance sampling, which will reduce the impact of policies which are further away from the current one.
%we can further leverage expected Sarsa, which be generalized to a multi-step as 
However, the importance sampling has drawback with high variance, which can be compensated with small step sizes but it will slow learning \cite{PrecupSS00}. In the next section we present a method that unifies active learning to select states and actions, and adaptive TD$(\lambda)$ to switch on/off backups of different stepsizes.

% -------------------------------------- our approach here --------------------------------------------------
\section{Active multi-step TD learning}
Our model consits of two pillars: active sample selection and adaptive multi-step TD learning. To speed up learning, we divide the time horizon into chunks and then select the most significant states. After we sample the states and actions, we propose an adaptive multi-step TD algorithm, which generalizes TD($\lambda$) by adaptively averaging backups of different $n$-step returns. %We first discuss how to select samples via active stepsize learning and time chunking. Then we introduce adaptive multi-step TD algorithm, which can be thought as a context-aware TD learning algorithm. %Finally, we propose the active actor-critic model, which extends the interval actor-critic, but with active stepsize selection.
\subsection{Active stepsize learning}
\begin{figure}[ht!]
%\vskip 0.2in
\centering
\includegraphics[clip, trim=20mm 65mm 20mm 35mm, width=8cm]{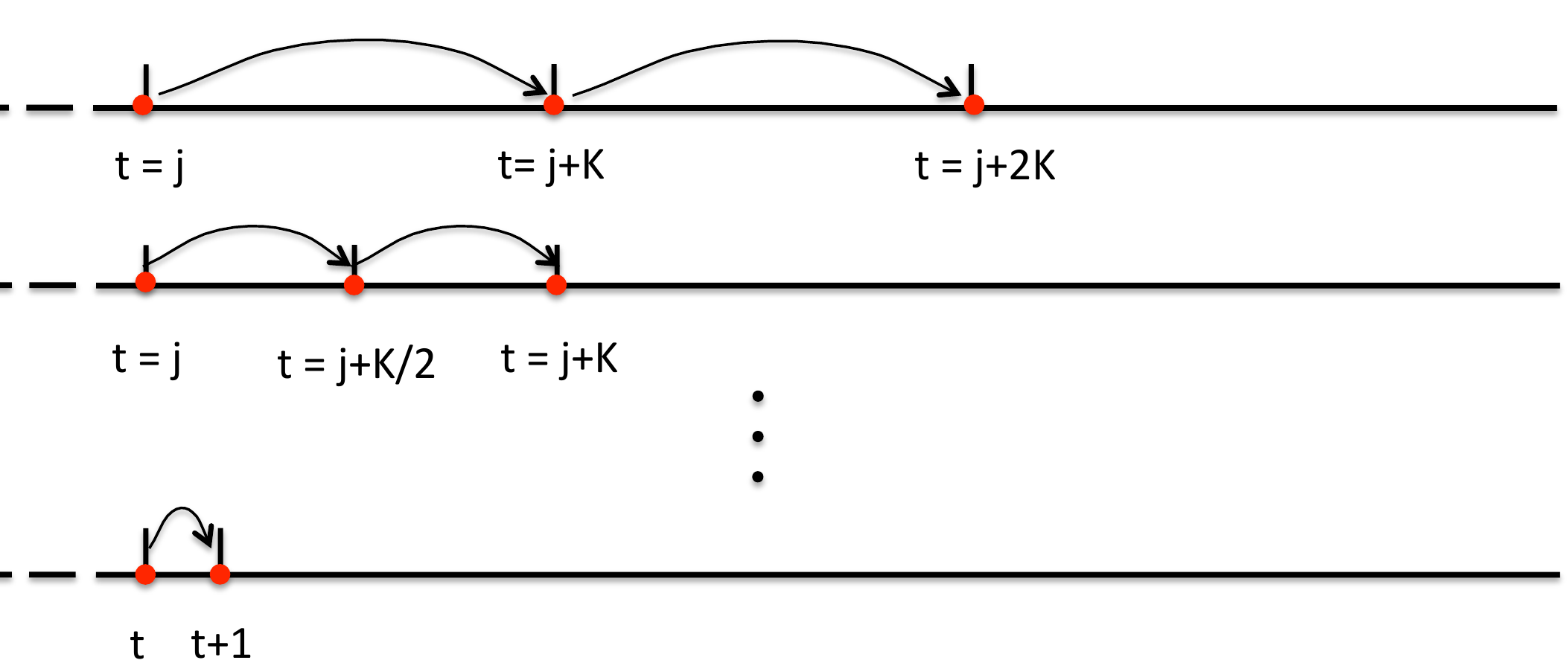}
\caption{The coarse-to-fine actor-critic approach to select the state (or the time step).}
\label{fig:ctfine}
%\vskip -0.2in
\end{figure}

For each trajectory $\tau=({s_0, a_0, s_1,..., s_T})$ with length $T$, We split the horizon into chunks in Figure \ref{fig:ctfine}. Specifically, we can define a interval size $K$, which equally splits the horizon into $[T/K]$ chunks. Inside of each interval, we can actively select the most significant states to speed up learning. According to Eq. \ref{eq:grad}, we define the most significant state $(s_t, a_t)$ either has higher temporal difference or policy uncertainty. Note that we do not consider random policy in this paper. Thus, given the actor and critic, we select the most important time $t$, by maximizing the following objective in each interval $t\in [j, j+K]$:
\begin{align} \label{eq:selection}
 \mathcal{L}(t; \theta) =  &   \delta_t(s_t)^2 + \beta  \textrm{H} \big( \textrm{log} \pi_{\theta} (a_t|s_t) \big) 
\end{align}
where $\textrm{H} \big(\textrm{log} \pi_{\theta} (a_t|s_t) \big)$ is the entropy, $\delta_t$ is the one step temporal difference in Eq. \ref{eq:deltat} and $\beta$ is the weight to balance the above two terms.

\begin{figure}[ht]
%\vskip 0.2in
\centering
\includegraphics[clip, trim=20mm 65mm 20mm 35mm, width=8cm]{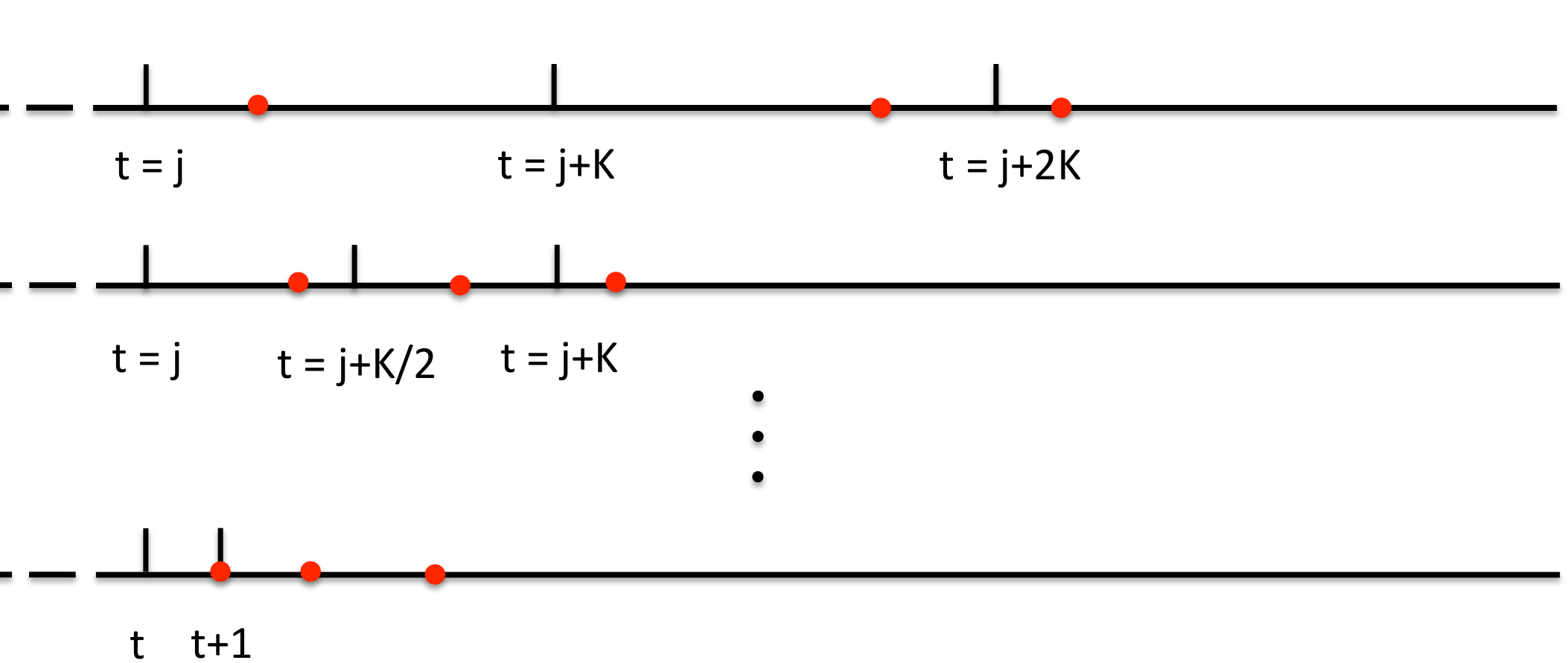}
\caption{The active actor-critic approach to select the state (or the time step) inside the intervals.}
\label{fig:adaptive}
%\vskip -0.2in
\end{figure}

%The difference between the right-hand and left-hand side of the Bellman equation, whether it is the one for the discounted reward setting  
For steps with higher $\delta_t$ and H($\cdot$) in Eq. \ref{eq:selection}, we should give them higher probability for exploration. So as in the active learning setting, we can always query actor-critic which step $t$ should be selected and further we can update our actor-critic models using TD learning. %If the TD error is positive, it suggests that the tendency to select should be strengthened for the future, whereas if the TD error is negative, it suggests the tendency should be weakened. Suppose actions are generated by the Gibbs softmax method:
Specifically, if the current state $s_t$ has higher uncertain policy, it should query actor-critic to improve its policy at the current state. Similarly, if the current value approximator has higher bias, it should query actor-critic to update its value function. Our objective looks at the time step $t$ which has higher policy uncertainty (which leads to higher variance) and higher bias. %With the coefficient $\beta$, we can balance the bias and variance. 
Moreover, by selecting the most significant time $(s_t, a_t)$, we can explore the action space well to converge to the optimal policy while $K\to1$ as TD(0). An example is shown in Fig. \ref{fig:adaptive}, where the red points are the steps selected by maximizing Eq. \ref{eq:selection} inside each interval.

%How can we implement the adaptive time step selection? Note that we split the each trajectory into chunks, and then we select the state $s_t$
%\subsection{Figures}
As for the discrete action space, we can take Eq. \ref{eq:selection} to select the state-action inside the chunk, which can implictly adjust stepsize between successive states and then optimze the policy. In the continouse action space, we can use Gaussian policy in the $A^2C$. However, it performs poorly compared to DDPG \cite{LillicrapHPHETS15}. Thus, we define another function for continous control below:
\begin{align} \label{eq:selection2}
 \mathcal{L}(t; \theta) =  \delta_t(s_t)^2 + \beta ( \nabla_{\theta}   \textrm{log}  \pi_{\theta}( a_t | s_t ) )^2
\end{align}
The difference between Eq. \ref{eq:selection} and Eq. \ref{eq:selection2} is that the later uses policy gradient for the continous action space. The high value in Eq. \ref{eq:selection2}, the high magnitude of gradient value in Eq. \ref{eq:grad}. In other words, we select the state-action pairs, which impacts the gradient most. %Analogously, if there is high entropy, it indicates the high action uncertainty, which implies high gradient improvement while we optimize the objective in Eq. \ref{eq:grad}.

Instead of updating actor-critic in every time step $t$, our strategy can select states inside intervals via active query to the current actor-critic architecture, which in turn implictly optimize the stepsize in the whole reinforcement learning framework. In the extreme case $K=1$, our model becomes the vanilla actor-critic method. Given the samples, we can use TD($n$) learning to update actor-critic model. Because we sample states, so the stepsize between successive states will be significant increased when $K>1$. One arising issue from TD($n$) learning may be the adverse effect from large $n$, which is even worse in continuous off-policy learning \cite{Harutyunyan2016QWO}. This phenomenon has also been mentioned early in the Tree-backup algorithm \cite{PrecupSS00}. 

%the interval actor-critic updates the model with chunk size $K$. In other words, we use TD($K$) learning to update actor-critic, where $K$ is the stepsize. As depicted in Fig. \ref{fig:ctfine}, the interval actor critic algorithm with time chunking only updates the model at steps marked by red points, which can significantly improve the training efficiency. However, it skips many steps, which contains important information such as rewards, so it may not converge to the optimal policy. 

%Thus, we gradually reduce the chunk size in the horizon, so that the chunked actor-critic algorithm can improve the sample efficiency while converging to the optimal policy. 
In the following section, we introduce our adaptive multi-step TD learning, which can avoid correction or importance sampling in the large stepsize case. 

\subsection{Adaptive multi-step TD learning}
Adaptive multi-step TD learning is a context-aware approach, which can decide whether to switch on or off future backups of different stepsizes. 
%combines the advantage of active learning and interval adaptive TD algorithm. It has an outer loop for active step selection (mentioned above) and an inner loop with multi-step TD as oracle. After selecting the most significant step $(s_t, a_t)$, we can update the policy and value model.
As mentioned before, given the interval length $K$, we can select the important states and actions based on Eq.\ref{eq:selection2}. Although the samples ($s_{t_{i}}$, $a_{t_{i}}$) at the different time $t_i$ is in order, the time difference ($t_{i+1} - t_{i} = 1$) between occurrence of successive data points (which are actively sampled) does not hold anymore if $K>1$. Fortunately, we can still use the average multi-step TD($n$) via Eq. \ref{eq:tdaverage} over the sampled data $\{ (s_{t_{i}}, a_{t_{i}} ) \}$. Here, we introduce the binary variables which can adaptively decide whether to truncate the long term returns or not while averaging backups of different step lengths. If the state and action are not consistent, it may indicate there is significant context change, then we can turn off the long term returns, and only use the near term backups to update our model. %, which is similar to TD(0). 
On the contrary, if the long term state and action is consisitent with the current environment, we can include it in our value estimation.
% n-step bootstrapping looks ahead to the next n rewards, states and actions, which generalizes Monte Carlo methods and one-step TD methods

The motivation we introduce the binary variables to TD$(\lambda)$ is to reduce its variance. In the early stage of learning,  $n$-step bootstrapping with large $n$ (as $n \to T$, it equals to the Monte Carlo return) can better fit the true value function. As the value function is better estimated, it is will be better if we can reduce its variance in the late stage of learning. Recall that:
\begin{align}\label{eq:motivation}
R_t^{n} = r_{t} + \gamma r_{t+1} + ...+ \gamma r_{t+n-1} + \gamma^n V(s_{t+n}) 
\end{align}
Then we can compute the variance of $R_t^{n}$ as:
\resizebox{.9\linewidth}{!}{
 \begin{minipage}{\linewidth}
\begin{align}\label{eq:variance}
Var( R_t^{n} ) =& Var[   R_t^{n-1} - \gamma^{n-1}V(s_{t+n-1}) + \gamma^{n-1} r_{t+n-1}  + \gamma^{n} V(s_{t+n})   ]\nonumber  \\
=& Var[   R_t^{n-1} - \gamma^{n-1}( r_{t+n-1} + V(s_{t+n}) -  V(s_{t+n-1}) )  ]  \nonumber \\
=& Var[R_t^{n-1}  ] +  \gamma^{2(n-1)} Var(  ( r_{t+n-1} + V(s_{t+n}) -  V(s_{t+n-1}) )  ) \nonumber \\
  & + 2Cov(R_t^{n-1},   ( r_{t+n-1} + V(s_{t+n}) -  V(s_{t+n-1}) ) ) \nonumber \\
\approx &  Var[R_t^{n-1}  ] +  \gamma^{2(n-1)} Var(  ( r_{t+n-1} + V(s_{t+n}) -  V(s_{t+n-1}) )  )  \nonumber \\
= &  Var[R_t^{n-1}  ] +  \gamma^{2(n-1)} Var(  \delta_t(n)  )
\end{align}

\end{minipage}
}

where we assume $Cov( R_t^n, R_t^{n-1} )  \approx  Var(R_t^{n-1}) $ considering the highly correlation between $R_t^n$ and $R_t^{n-1}$ in successive sequences. So that we have $Cov(R_t^{n-1},   ( r_{t+n-1} + V(s_{t+n}) -  V(s_{t+n-1}) ) ) \approx 0$ as in \cite{Kearns00}. As $n$ increasing, the variance of the $n$-step return $R_t^{n}$ increases correspondingly, which is contributed from $Var (\delta_t (n) )$. In addition, $\delta_t(n)$ can be thought as the advantage over the current state. If its sign changes, it may lead to high variance. Thus, if $Var(\delta_t(n) )$ changes rapidly, we can truncate the future backups to reduce the variance of TD$(\lambda)$. 

Hence, we extend Eq. \ref{eq:tdaverage} by introducing binary variables to turn on or off different steps while we compute the average return at $t_i$ 
\begin{align}\label{eq:tdfinal}
R_{t_i}^{avg} = \frac{ \lambda_1 b_1 R_{t_{i}}^{n_{1}}  + \lambda_2 b_2 R_{t_i}^{n_{2}} + ...  + \lambda_l b_l R_{t_i}^{n_{l}}} {   \lambda_1 b_1 + \lambda_2 b_2 + ... + \lambda_l  b_l }
\end{align}
where $R_{t_i}^{n_{l}}$ is defined via Eq. \ref{eq:esarsa}, and $b_i, i=\{1,2, ..., l\}$ is the context-aware binary variable, which will be learned based on the consisitency between states and actions. If ${b_l=0}$, then its branch backup from $R_{t_i}^{n_{l}}$ will be turned off. Otherwise, it will include $R_{t_i}^{n_{l}}$ in the average of backups above. 

In our model, we train a binary classifier $f(s, a)$ to model context change. Especially, given $(s_{t_i}, a_{t_i})$, we use the sign of the advantage $Q(s_{t_i}, a_{t_i}) - V(s_{t_i})$ as the groundtruth, then we can learn a binary classifier $y= f(s, a)$. So for the future environment $(s_{n_l}, a_{n_l})$, we define $b_l$ as follows
\begin{align}
b_l  = [f(s_{t_i}, a_{t_i}) == f(s_{n_l}, a_{n_l})]
\end{align}
The purpose we introduce $b_l$ for TD($n_l$) is to capture the context consistence while we average the different stepsize returns. So our model is a context-aware approach, which can automatically turn on/off certain backups if it detects significantly environmental changes. In addition, we can add more information, such as stepsize and action difference, to the context except states and actions while learning the binary classifier.

\subsection{Algorithms}
Assume that we have $N$ trajectories ${ \{\tau_i \} _{i=1}^N}$, where $\tau_i =({ s_0, a_0, s_1,..., s_{T_i}  })$. Further, we define a list of $M$ intervals $\mathcal{K}=\{K_1, K_2, ..., K_M\}$, where each $K_m$ specifies the interval size. If we take a coarse to fine scale in the horizon, then intervals satisfy $T>K_1\geq...\geq K_M=1$, shown in Fig. \ref{fig:ctfine}. Moreover, the step distance between samples in successive intervals will vary from time to time, shown in Fig. \ref{fig:adaptive}. On the contrary, we can also take a fine to coarse approach, by setting $1\leq K_1\leq...\leq K_M <T$. Of course, we can set a fixed $K_m$ for any $m$ over all these trajectories. Note that $K_m$ for $m \in [1, M]$ can be any integer as long as $K_i \in [1, T]$. In addition, we can calculate the number of trajectories $N_i$ assigned to each interval, where $N_i = [N/M]$ if the trajectories are equally distributed over the total $M$ intervals. Our active multi-step TD in Algorithm \ref{alg:aac} comprises two loops: the loop for state selection and the loop to update actor-critic model using adpative multi-step TD learning. In the active selection stage, we do the inference over the steps $t \in [j, j+K]$, where $K \in \mathcal{K}$, and select $(s_t, a_t )$ by maximizing Eq. \ref{eq:selection2}. In the learning stage, we can consider two situatioins: on-policy and off-policy learning. In the on-policy case, we can immediately update the model with TD(0). As for the off-policy, we can put the sampled data into buffer, and then we can update actor-critic with adpative multi-step TD by sampling the data from the buffer. Note that we can use any state of the art TD algorithm in our context-aware framework to improve the performance. Note that $r_t$ in Algorithm \ref{alg:aac} is the accumulated reward over the past steps under the behavior policy.

\begin{algorithm}[tb]
   \caption{Active multi-step TD algorithm}
   \label{alg:aac}
\begin{algorithmic}
   \STATE Initialize critic network $Q(s, a | \theta^{Q})$ and actor $\mu(s | \theta^{\mu})$ with weights $\theta^{Q}$ and $\theta^{\mu}$
   \STATE Initialize the binary classifier $f(s,a)$
   \STATE Initialize $M$ intervals $[K_1, K_2, ..., K_M]$ 
   \STATE Initialize the total number of episodes $N$ and the replay buffer $\textbf{\emph{R}}$
   \STATE Initialize the coefficient $\beta$ in Eq. \ref{eq:selection2}
   \STATE Initialize the number of lookahead backups $l$
   \STATE Calculate the number of episodes for each interval $[N/M]$
   \FOR{episode = 1 {\bfseries to} $N$}
       \STATE Compute the interval size $K = K_{episode\%M}$ 
       \STATE $D$ = [ ] // samples from the trajectory
       \STATE Receive initial observation state $s_0$ from the environment 
       \FOR{$t=0$ {\bfseries to} $T$}
           \STATE Select action according to $a_t = \mu(s_t | \theta^{\mu})$
           \STATE Execute action $a_t$ and receive reward $r_t$, $done$, and further observe new state $s_{t+1}$ 
           %\STATE Inference the value function $V(s_t)$ and the time difference $\delta_t = \big(  r_t  +  \gamma V(s_{t+1})  \big)  - V(s_t)$
           %\STATE Calculate the entropy $\textrm{H} \big(\textrm{log} \pi_{\theta} (a_t|s_t) \big)$ in Eq. \ref{eq:selection}
           \STATE Compute the selection function $ \mathcal{L}(t; \theta)$ in Eq. \ref{eq:selection2}       
           \IF{$t \%K ==0$}
             \STATE Select the most significant state $(s_t, a_t)$ %inside chunk $K$ which maximizes $ \mathcal{L}(t; \theta)$ in Eq. \ref{eq:selection2}
             %\STATE Call actor-critic algorithm to update weights $\theta^{Q}$ and $\theta^{\mu}$
             \STATE Add the sample $(s_t, a_t, r_t, t, s_{t+1})$ into $D$
             %\STATE Store state information $(s_t, a_t, r_t, t )$ to $\textbf{\emph{R}}$ 
           \ENDIF
       \ENDFOR
       \FOR{ $t$  =1 to length($D$) }
           \STATE Push continuous $l$ tuples of $(s_t, a_t, r_t, t, s_{t+1})$ into $\textbf{\emph{R}}$
       \ENDFOR
       \STATE Update the actor and critic via Algorithm \ref{alg:atd}
   \ENDFOR
   \STATE Return parameters $\theta =  \{ \theta^{Q}, \theta^{\mu} \}$.
\end{algorithmic}
\end{algorithm}

\begin{algorithm}[tb]
   \caption{Adaptive multi-step TD algorithm}
   %\KwResult{Write here the result}
   %\SetKwInOut{Input}{Input}\SetKwInOut{Output}{Output}
   %\Input{Write here the input}
   %\Output{Write here the output}
   %\BlankLine
   \label{alg:atd}
\begin{algorithmic}
   \STATE{\bf Input:} critic network $Q(s, a | \theta^{Q})$ and actor $\mu(s | \theta^{\mu})$, and their target networks
   \STATE {\bf Input:} the binary classifier $f(s,a)$
   \STATE {\bf Input:} the number of lookahead backups $l$
   \STATE{\bf Input:} the number of iterations $Iter$ 
   \STATE{\bf Input:} the batchsize 
   \FOR{i = 1 {\bfseries to} $Iter$}
       \STATE Sample a batchsize of $l$ continuous tuples from $\textbf{\emph{R}}$: $(s_{t_i}, a_{t_i}, r_{t_i}, t_i, s_{t_i + 1} )$, ..., $(s_{t_{i} +n_l }, a_{t_{i} +n_l }, r_{t_{i}+n_l}, t_{{i}+l}, s_{t_{i} +n_l} )$ 
       \STATE Infer the binary variables $b_1$, ..., $b_l$ via $f(s,a)$
       \STATE Compute the average return at ${t_i}$ via Eq. \ref{eq:tdfinal}
       \STATE Compute the advantage and its sign $y_{t_i}$, \\
       \STATE Prepare the training data $\{ (s_{t_i}, a_{t_i}), y_{t_i} \}$
       \STATE Update critic $Q(s, a | \theta^{Q})$ 
       \STATE Update actor $\mu(s | \theta^{\mu})$
       \STATE Update target networks
       \STATE Update the classifier $f(s,a)$ with $\{ (s_{t_i}, a_{t_i}), y_{t_i} \}$
   \ENDFOR
   \STATE Return parameters $\theta =  \{ \theta^{Q}, \theta^{\mu} \}$ and the classifier $f(s,a)$.
\end{algorithmic}
\end{algorithm}

%\subsection{Comparisoin to other algorithms}
%
%Our active multi-step TD method selects the significant state-action pair, which implictly learns the stepsize in time difference approach. In our algorithm \ref{alg:aac}, our approach use active learning, which actively search $(s_t, a_t)$ and then update policy. Furthermore, we take advantage of replay buffer, and use two critics as TD3 \cite{FujimotoHM18} to handle the overestimated bias.  

% Note use of \abovespace and \belowspace to get reasonable spacing
% above and below tabular lines.

\begin{figure*}[t!]
\begin{tabular}{ccc}
\includegraphics[trim=15mm 8.8mm 20mm 15mm, clip, width=5.4cm]{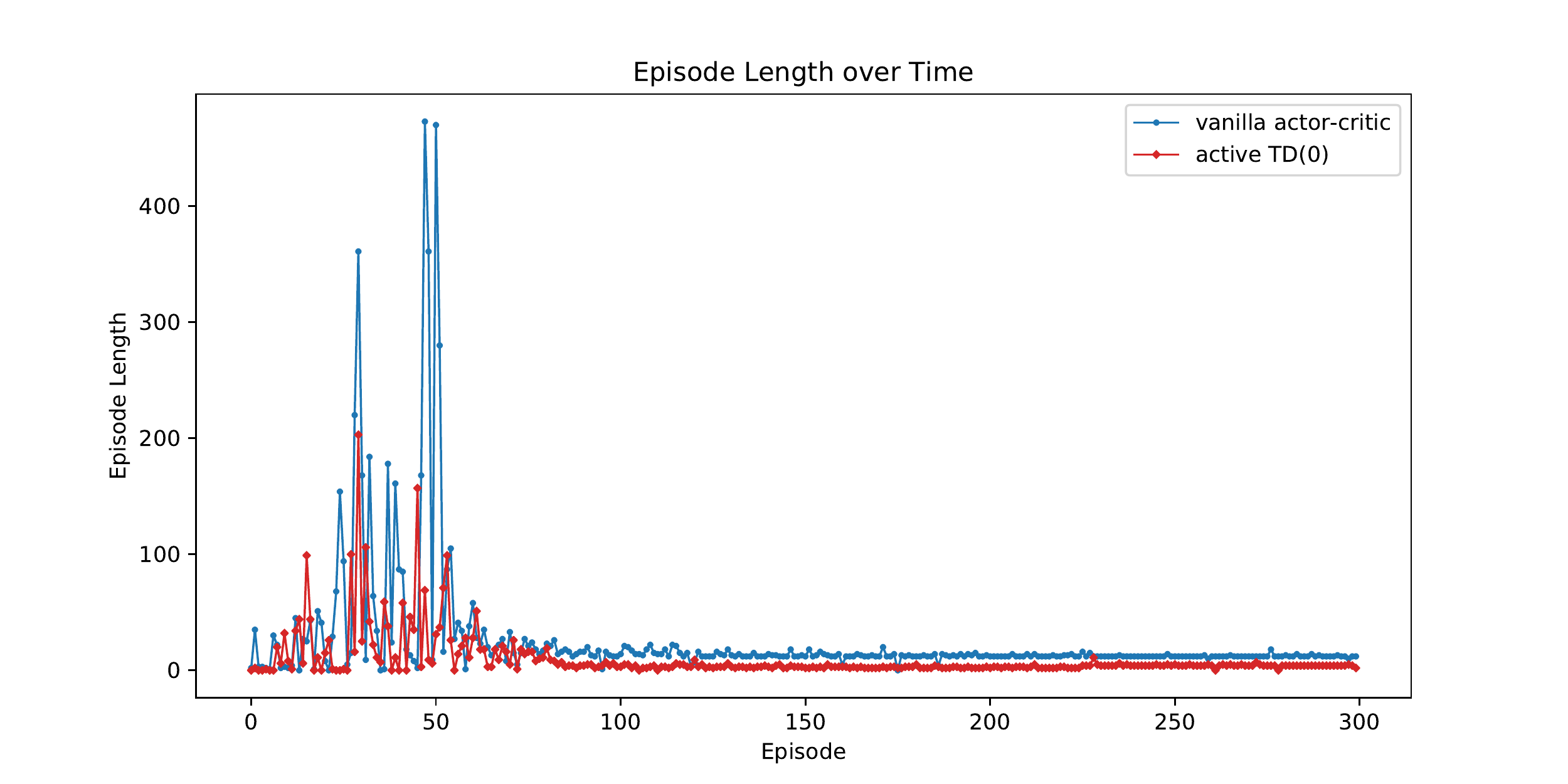} &
\includegraphics[trim=15mm 8.8mm 20mm 15mm, clip, width=5.4cm]{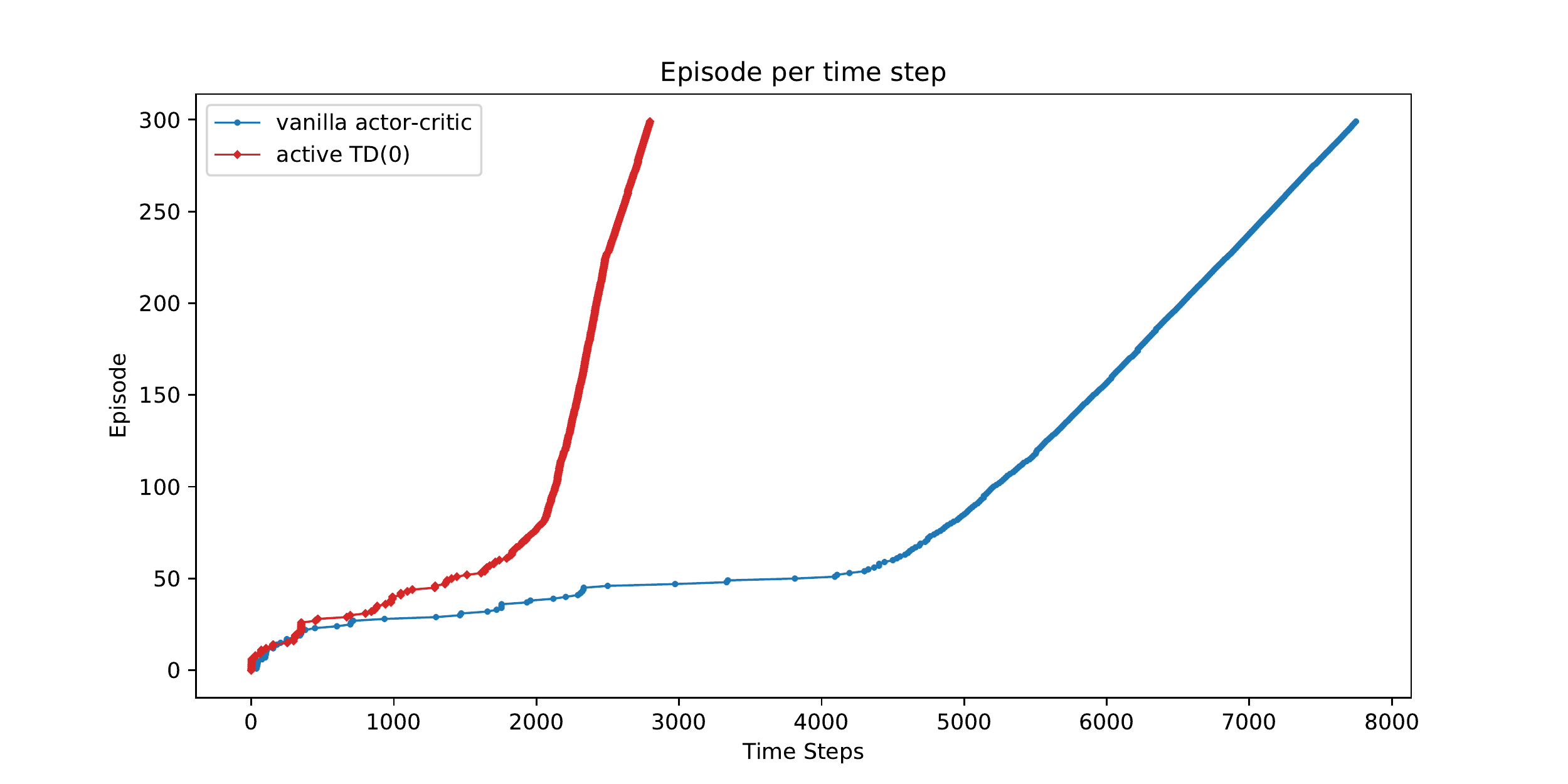} &
\includegraphics[trim=15mm 8.8mm 20mm 15mm, clip, width=5.4cm]{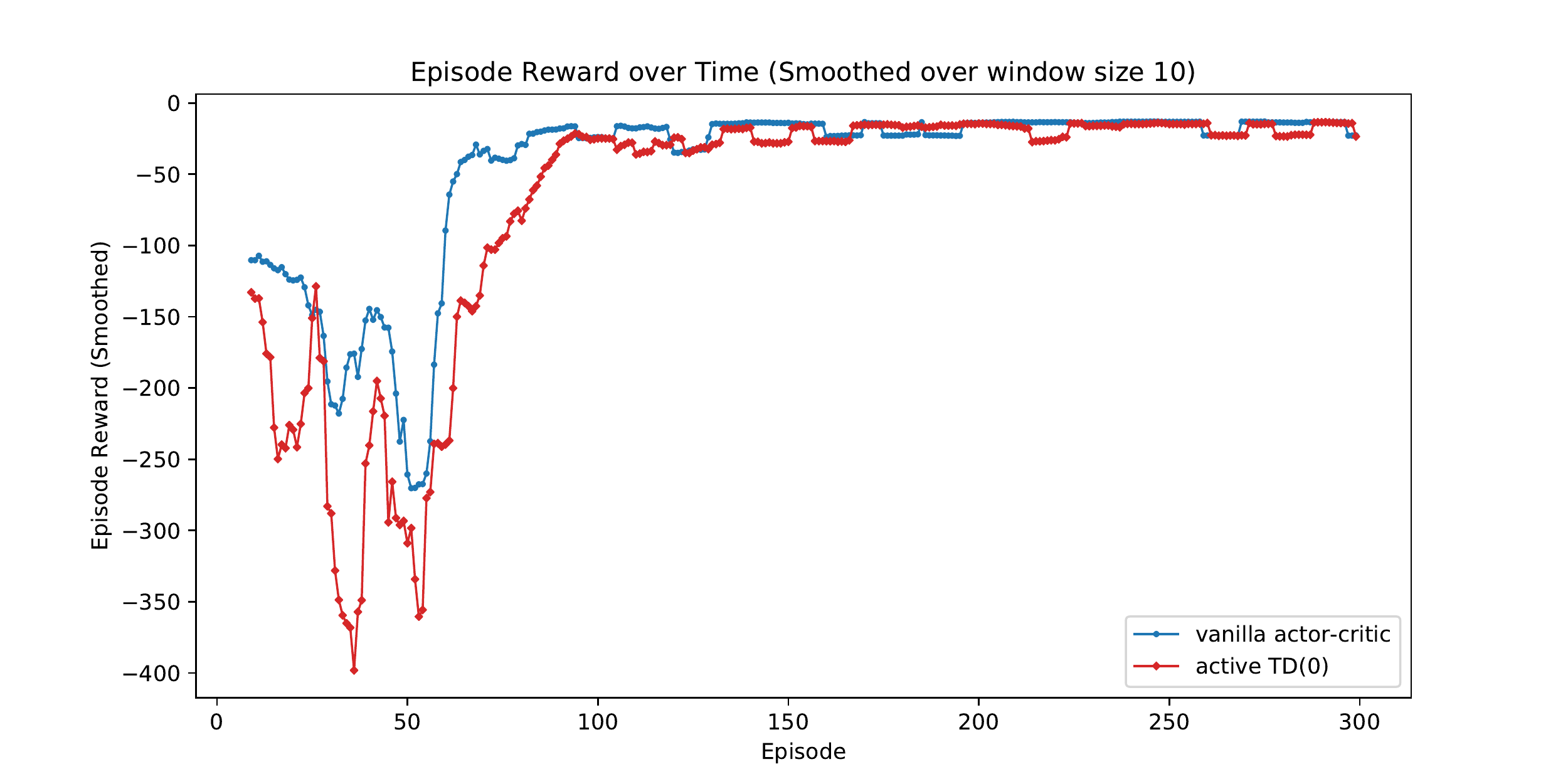} \\
(a) & (b) & (c)
\end{tabular}
\caption{The results on the Cliff Walking problem. (a) and (b) show how many episodes are required to reach the maximum return, which demonstrates that our method with active selection can use the training data more effectively. (c) indicates how the smoothed (window size 10) reward changes while increasing the number of episodes in the learning process. Our active TD shows much stable results with more episodes.}
\label{Fig:cliffwalk}
\end{figure*}

% ---------------------------------------------- experimental results ------------------------------------------
\section{Experiments}
In this section, we evaluate our method on both descrete and continous action space. In particular, we test whether the active selection works or not without including the context-aware strategy. In addition, we also combines both active learning and off-policy adaptive multi-step TD in algoirthm \ref{alg:aac} to verify whether the context-aware strategy contributes to boost the overall return.

In the descrete action space, we mainly test whether the active selection contributes or not in our active TD algorithm. We use a simple TD(0) learning with actor and critic networks, and sample the action based on the actor network. The deep achitecture for both actor and critic uses a hidden layer with 20 dimensions. The batch size is set 100, learning rate is 0.001 for actor and 0.01 for critic respectively, and $\beta=0.1$. While we update actor and critic, namely TD(0), we do not consider returns of different stepsizes. So we set $K_m = 1$ for all $m \in [1, M]$ and $l=1$ in Eq. \ref{eq:tdfinal}, referring Algorithm \ref{alg:atd} for more details. Both target networks are updated with $\tau = 0.001$ if not specified. We compared our approach with other baselines, such as active-critic, REINFORCE and DQN. 

\begin{figure*}[t!]
\begin{tabular}{ccc}
\includegraphics[trim=13mm 8.8mm 20mm 15mm, clip, width=5.4cm]{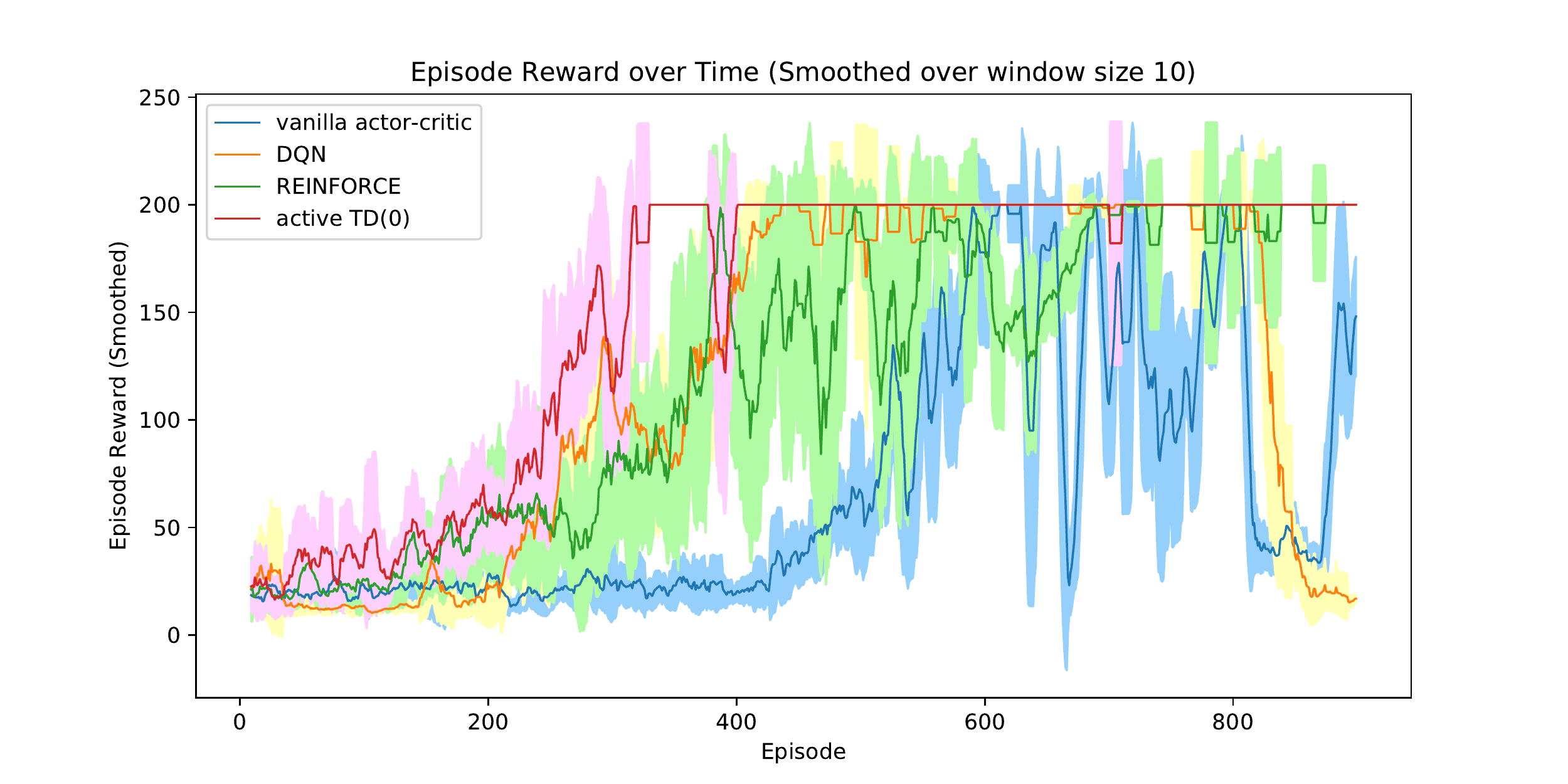} &
\includegraphics[trim=13mm 8.8mm 20mm 15mm, clip, width=5.4cm]{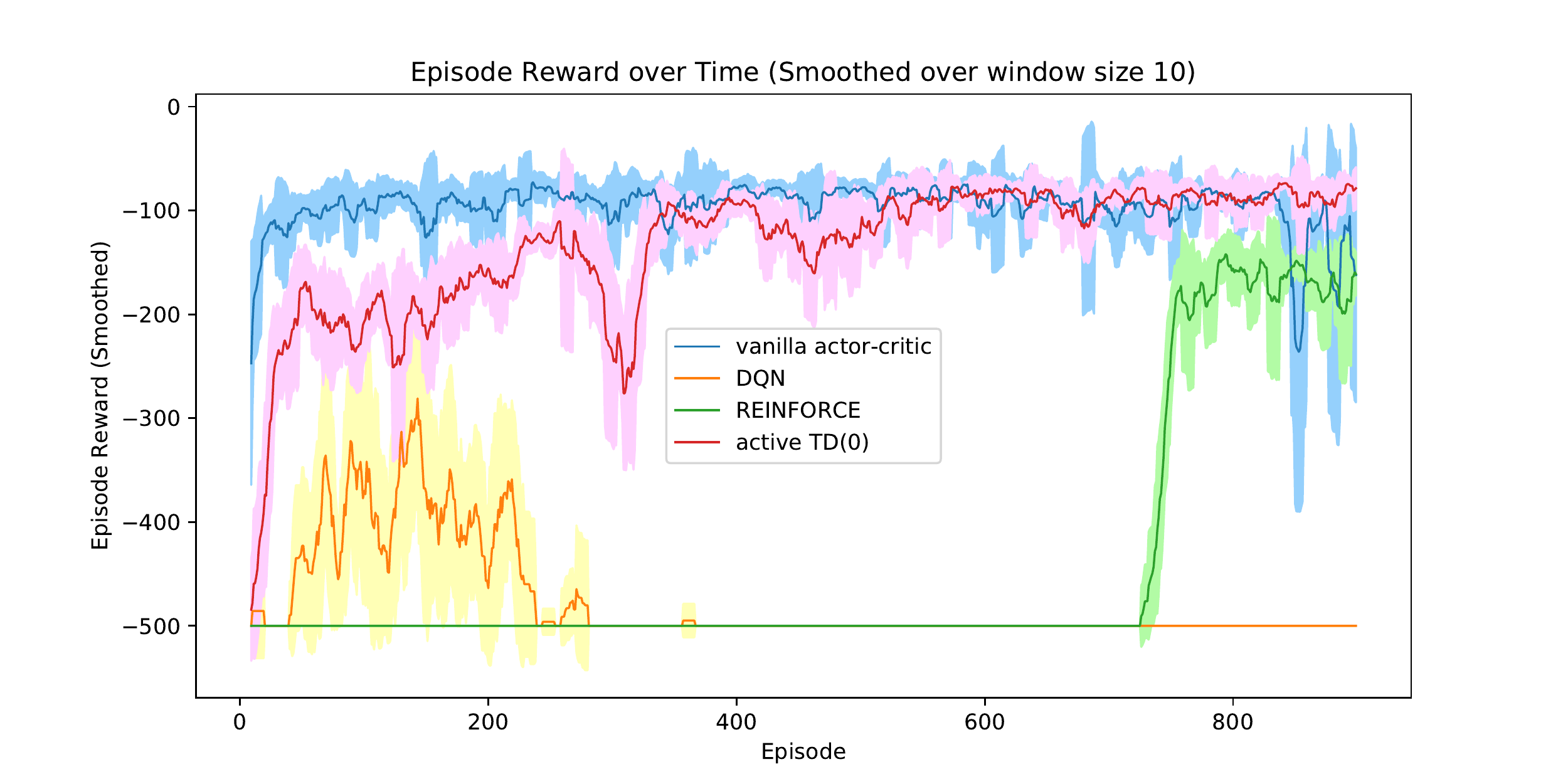} &
\includegraphics[trim=13mm 8.8mm 20mm 15mm, clip, width=5.4cm]{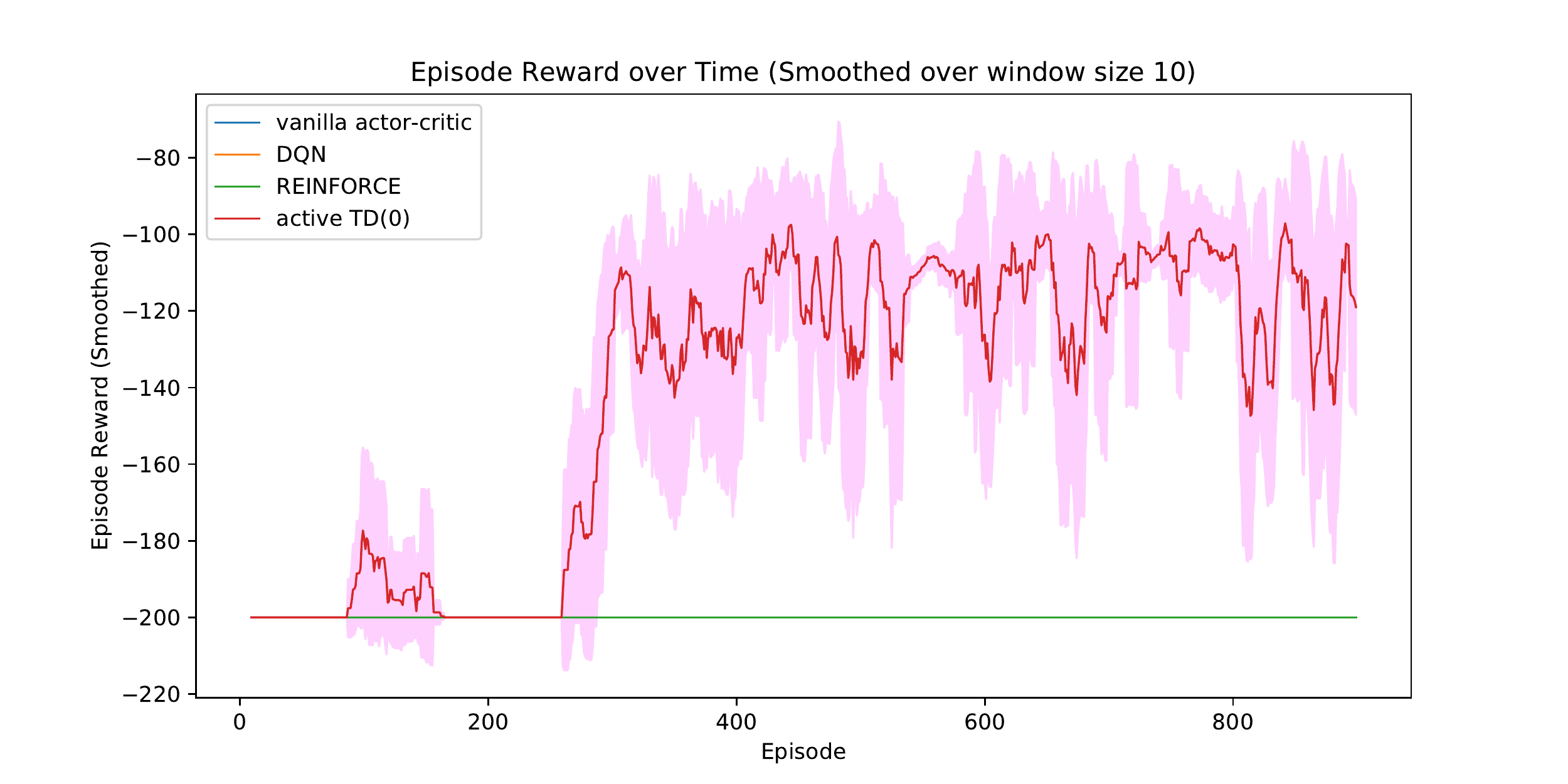} \\
(a) Cartpole & (b) Acrobot & (c) MoutainCar
\end{tabular}
\caption{The result (smoothed over window size 10) on the Cartpole, Acrobot and MoutainCar environments. It demonstrates that our active sample selection can boost performance.}
\label{Fig:descrete}
\end{figure*}

The result on cliff walking environment is shown in Figure \ref{Fig:cliffwalk}. Compared to vanilla actor-critic, our model can leverage the episodes more efficiently and achieve optimal state with short steps, shown in Figure \ref{Fig:cliffwalk}(a). In addition, we test our approach on another three classical environments, and the learning curves shown in Figure \ref{Fig:descrete} indicate our approach matches or outperform the other three baselines. Fig. \ref{Fig:descrete}(a) shows the comparison on Cartpole enviroment. Our simple active TD(0) has better average return, compared to other methods. Similar result is observed in MoutainCar environment. In the Acrobot enviroment, vanilla actor-critic is better at the beginning episode. But our method yield much better and stable returns in the late episodes.

In the continous action space, we mainly test adaptive multi-step TD with the interval size $K_m=4$ fixed for all $m \in [1, M]$ and measure its performance on a suite of different control tasks. Without other specification, we use the same parameters below for all environments. The deep achitecture for both actor and critic uses the same networks as TD3 \cite{FujimotoHM18}, with hidden layers [400, 300, 300]. Note that the actor adds the noise $\mathcal{N}(0, 0.1)$ to its action space to enhance exploration and the critic network has two Q-functions as TD3. In addition, we also use target networks (for both actor and critic) to improve the performance as in DDPG and TD3. The target policy is smoothed by adding Gaussian noise $\mathcal{N}(0, 0.2)$ as in TD3. The classifier uses the same network structure as critic, but with binary (softmax) output. The number of lookaheads $l=3$ to incorporate different backups. For $n_1$, we use the minimum of the dual Q-function as TD3 to get the target value, while for $n_2$ and $n_3$, we use the average of the dual Q-value in Eq. \ref{eq:esarsa}. The weights $\lambda_i$ over TD($n_i$) for $i=\{1,2,3\}$ decay exponentially with the base $\frac{1}{2}$ as in TD($\lambda$). Both target networks are updated with $\tau = 0.005$. In addtion, the off-policy algorithm uses the replay buffer with size $10^6$ for all experiments as TD3 did.%As for BipedalWalker, We use the list of intervals [6, 6, 3, 3] with episode 900 to actively select states and actions, and the rate to update target networks $\tau = 0.001$. 

\begin{figure*}[t!]
\begin{tabular}{ccc}
\includegraphics[trim=13mm 8.8mm 20mm 15mm, clip, width=5.4cm]{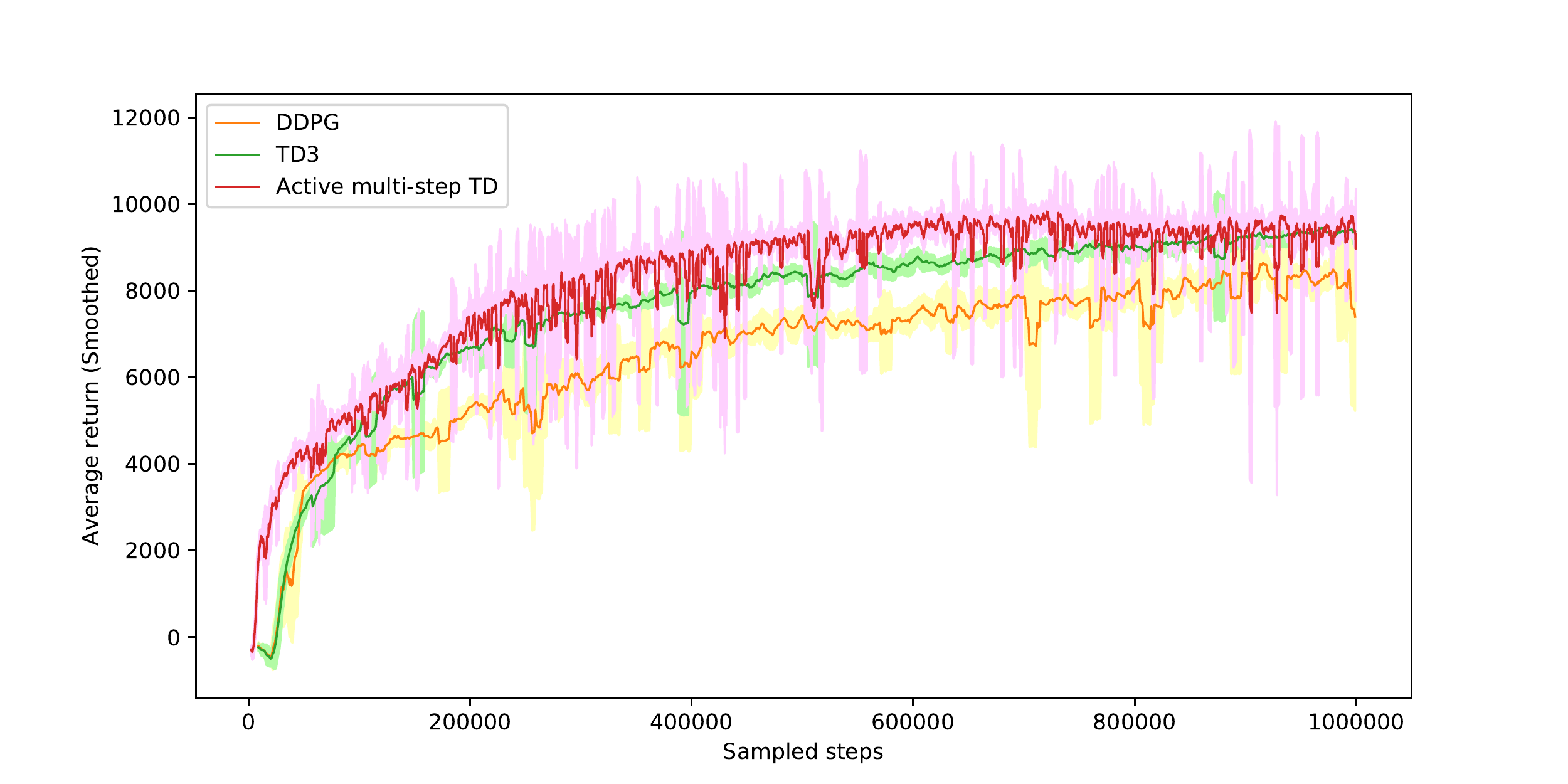} &
\includegraphics[trim=13mm 8.8mm 20mm 15mm, clip, width=5.4cm]{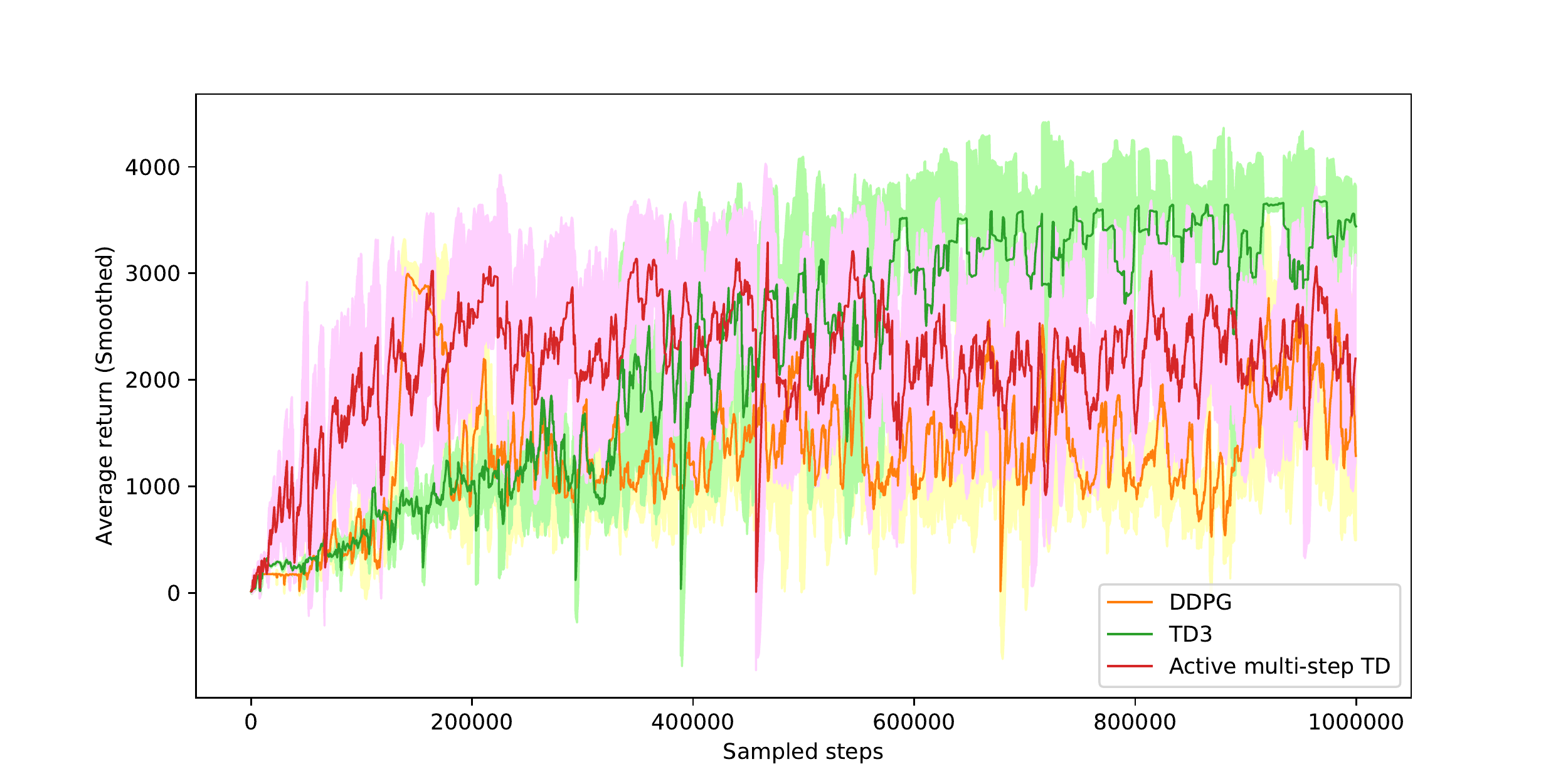} &
\includegraphics[trim=13mm 8.8mm 20mm 15mm, clip, width=5.4cm]{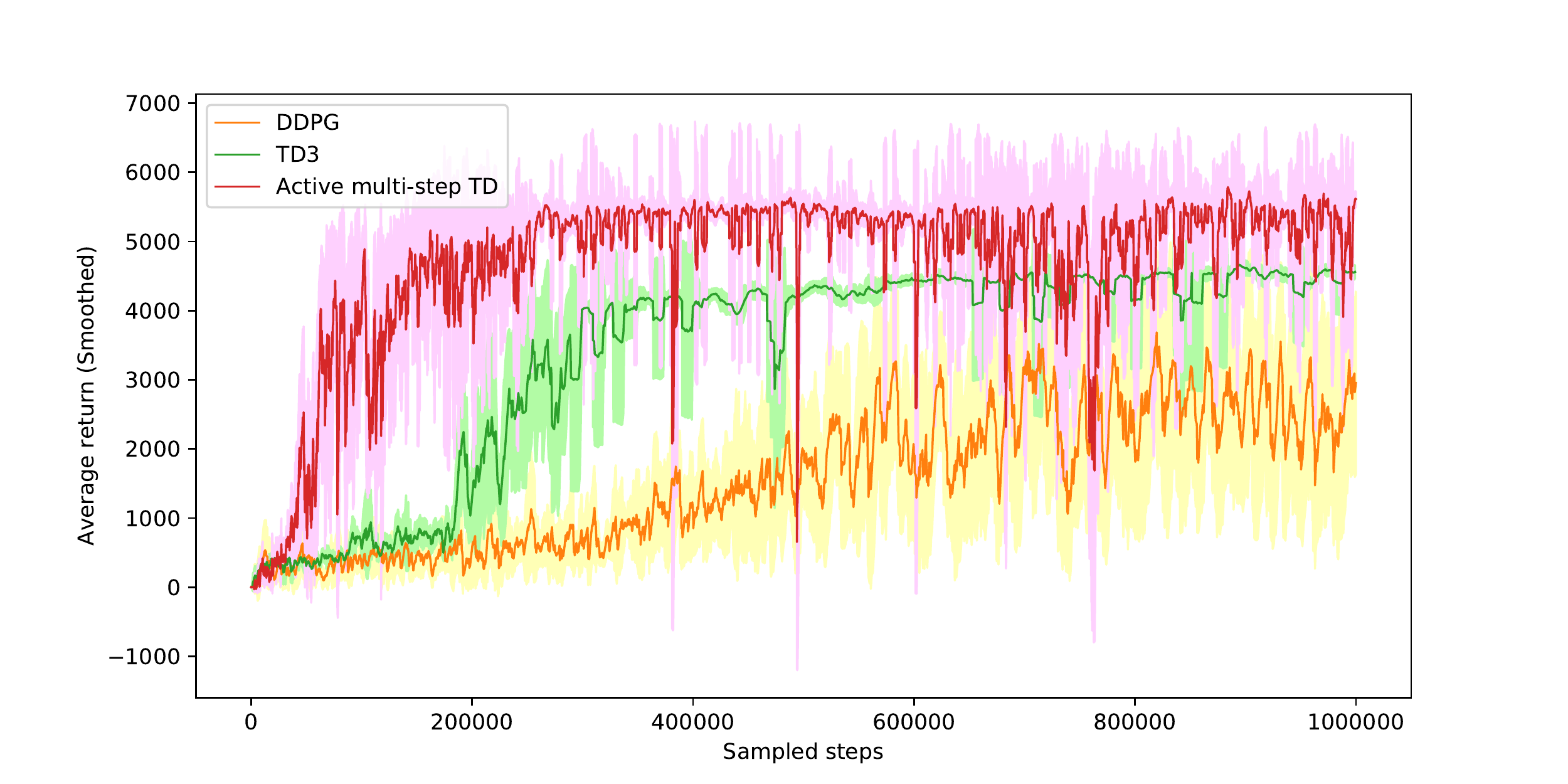} \\
(a) Halfcheetah & (b) Hopper & (c) Walker2d
\end{tabular}
\caption{The Learning curves with exploration noise on the Halfcheetah, Hopper and Walker2d environments. The shaded region represents the standard deviation of the average evaluation over nearby windows with size 10. Our active multi-step TD algorithm yields significantly better results, compared to TD3 and DDPG.}
\label{Fig:MuJoCo1}
\end{figure*}

\begin{figure*}[t!]
\begin{tabular}{ccc}
\includegraphics[trim=13mm 8.8mm 20mm 15mm, clip, width=5.4cm]{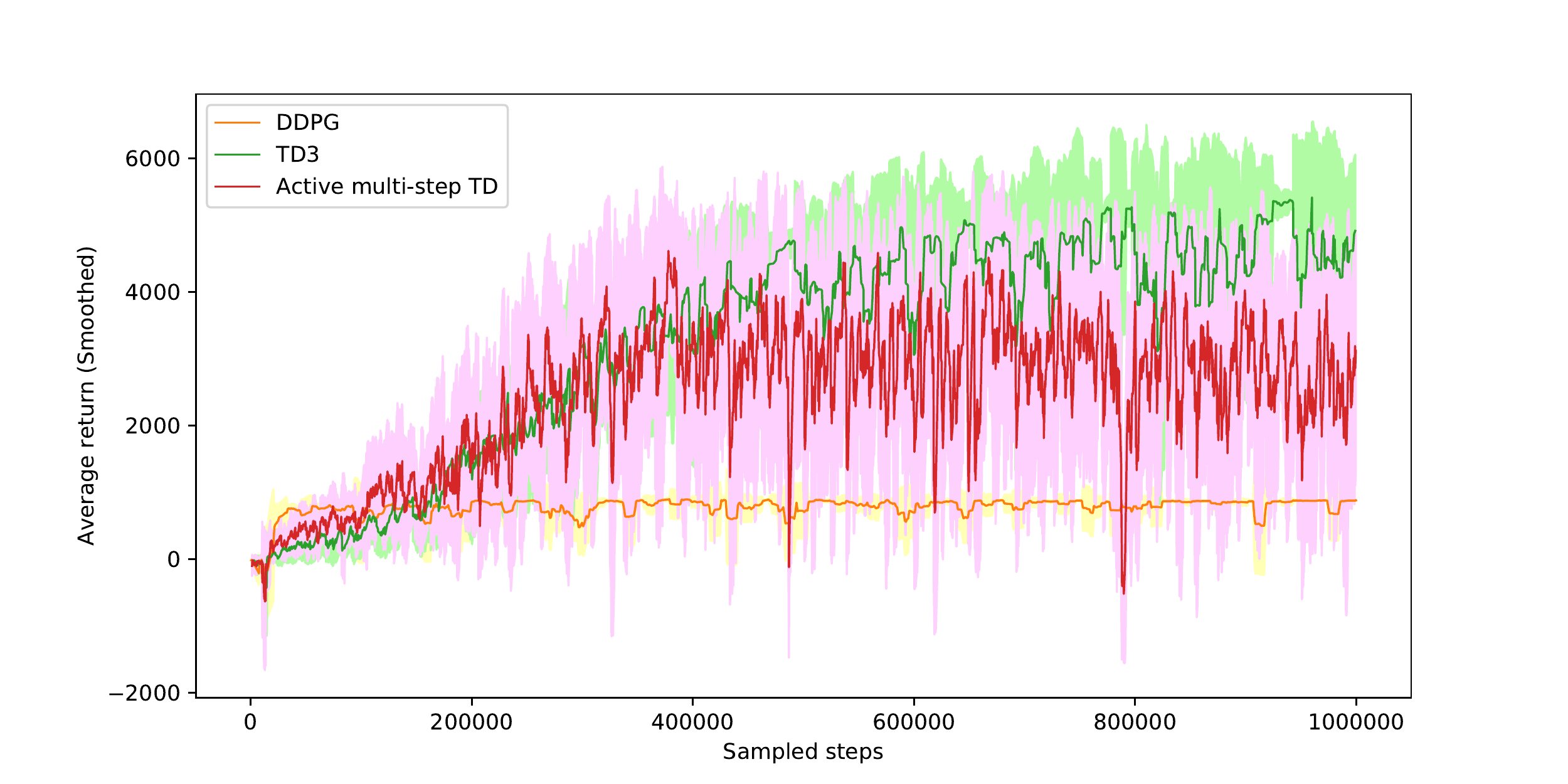} &
\includegraphics[trim=13mm 8.8mm 20mm 15mm, clip, width=5.4cm]{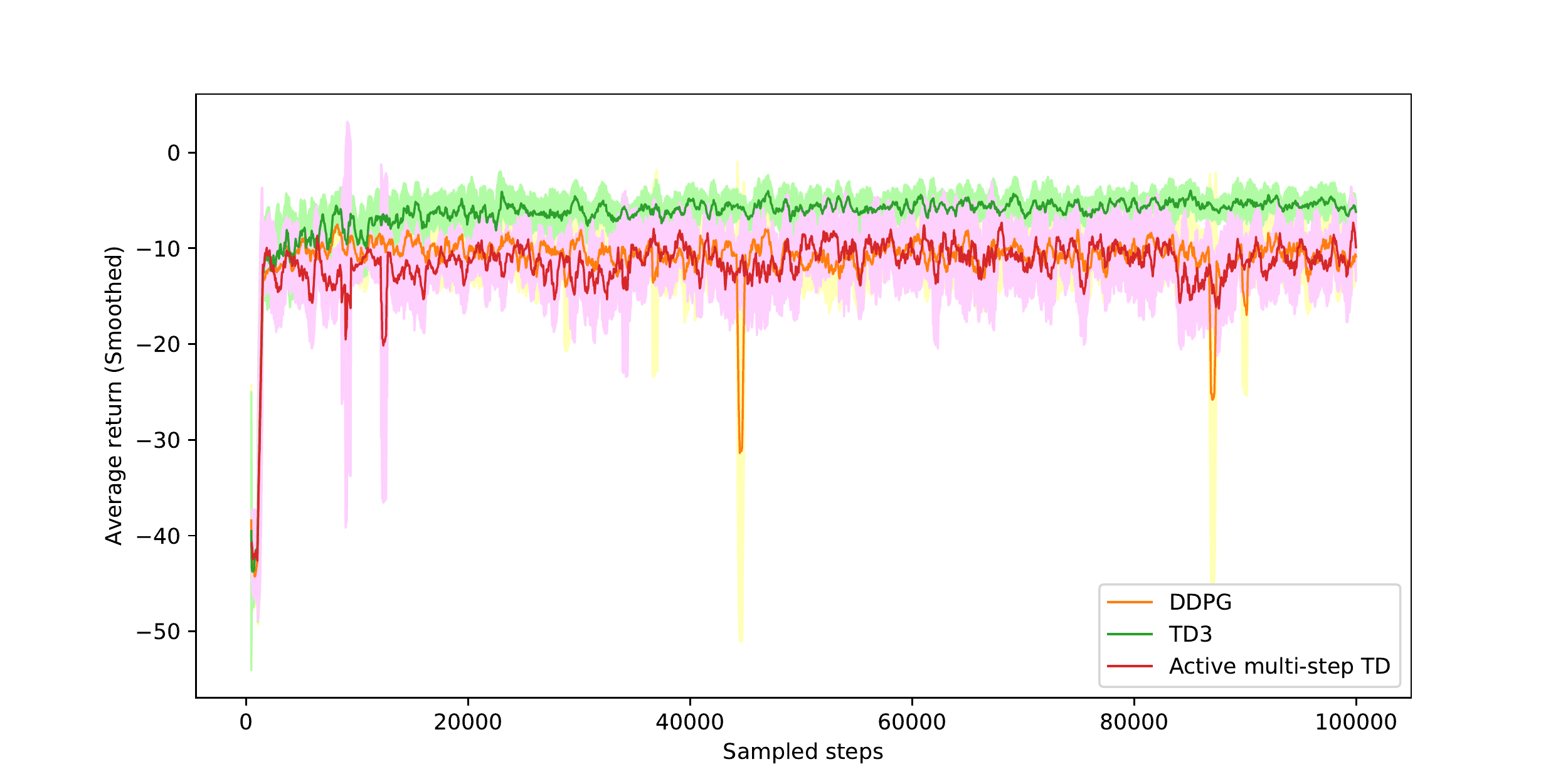} &
\includegraphics[trim=13mm 8.8mm 20mm 15mm, clip, width=5.4cm]{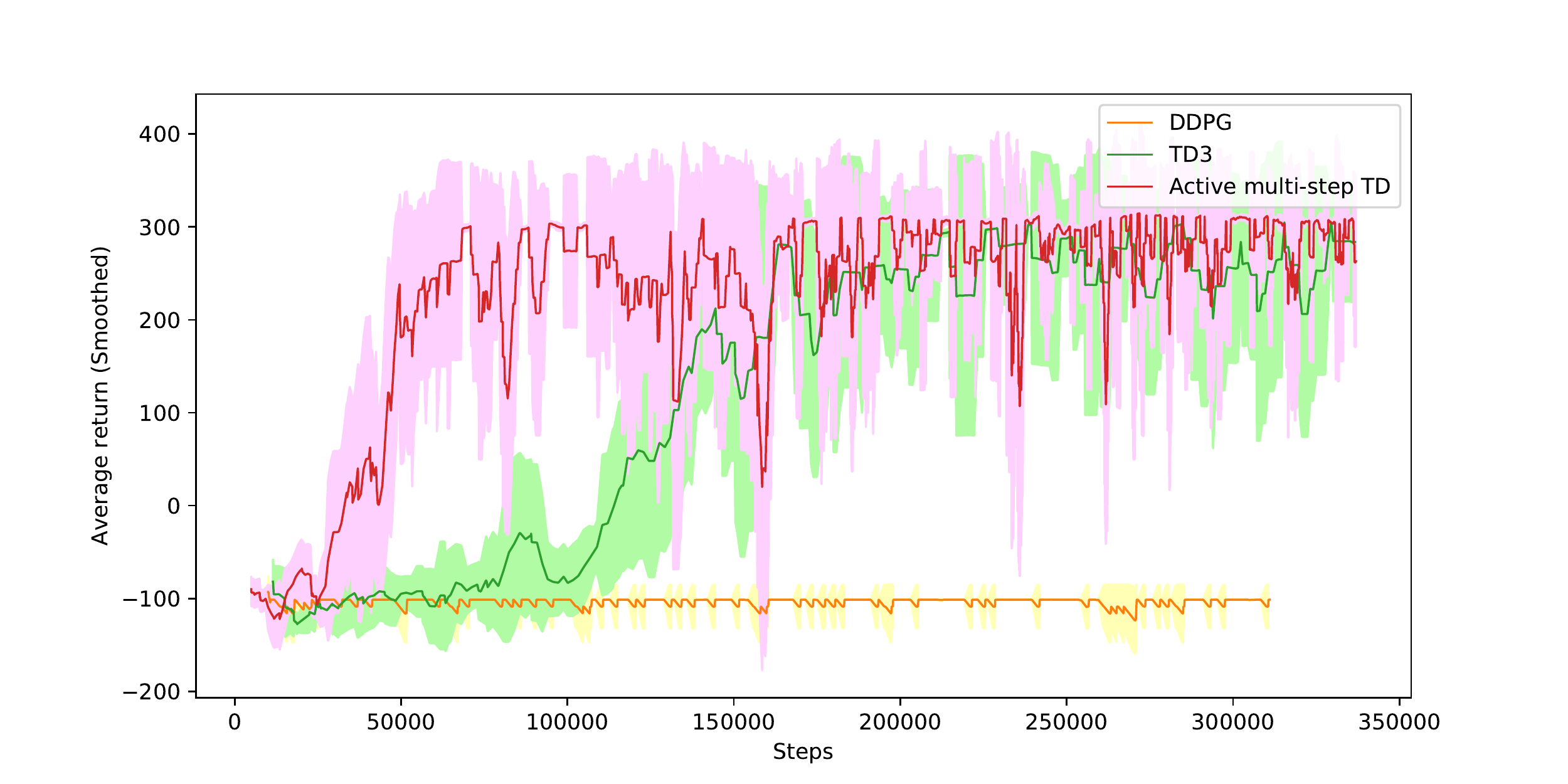} \\
(a) Ant & (b) Reacher & (c) BipedalWalker
\end{tabular}
\caption{Learning curves for the OpenAI gym continuous control tasks with exploration noise. The shaded region represents the standard deviation of the
average evaluation over nearby windows with size 10. It shows that our approch on BipedalWalker can more effeciently use the training samples and yield higher returns.}
\label{Fig:MuJoCo2}
\end{figure*}

The earning curves with exploration noise are shown in Figures \ref{Fig:MuJoCo1} and \ref{Fig:MuJoCo2}. It demonstrates that our approach can yield competitive results, compared to TD3 and DDPG. Specifically, our active multi-step TD approach outperforms all other algorithms on HalfCheetah, Walker2d and BipedalWalker in both final performance and learning speed across all tasks. The quatitative results over 5 trials are presented in Table \ref{Tab:tab1}. Except Reacher and BipedalWalker, we learn the model until to 1 million samples and then evaluate the reward. We repeat this with 5 trials and then we can calcualte the average return. For Reacher task, it converges fast, so we only sample 100k steps. Similarly, we use 350k samples in BipedalWalker environment. It shows that our approach yields significantly better results on HalfCheetah, Walker2d and BipedalWalker. Note that we only try the fix interval size $K_m=4$. By varying the interval size, our approach can easily yield comparable or even better results on the other three MuJoCo environments.

\begin{table*}[!htbp]
\centering
\scalebox{0.73}{
\begin{tabular}{*8c}
\toprule
\multirow{2}{*}{Methods} &  \multicolumn{7}{c}{Environments} \\
{}   & HalfCheetah & Hopper  & Walker2d   & Ant &  Reacher & InvPendulum & BipedalWalker \\
\midrule
DDPG   &  8577.29  & 2020.46   & 1843.85 & 1005.30 & -6.51 & 1000  & $-101.40 \pm 0.4$\\
TD3  &  $9636.95\pm859.06$ & $ \bf{3564.07\pm114.74} $  & $4682.82\pm539.64$  & $\bf{4373.44\pm1000.33}$ & $ \bf{-3.6\pm0.56 }$ & $1000\pm0.0$  & $255.6\pm74.2$\\
Active Multi-step TD   &  $ \bf{9836.5\pm611.10} $ &  $3063.05 \pm 357.21$    & $ \bf{5092.57\pm270.71} $  & $3135.67\pm325.86$ & $-10.21 \pm -1.64$ & $ \bf{1000 \pm 0.0} $ & $ \bf{275.03\pm 98.1} $\\
\bottomrule
\end{tabular}}
\caption{Comparison of Max Average Return over 5 trials of 1 million samples, except Reacher and BipedalWalker. The maximum value is marked bold for each task. $\pm$ corresponds to a single
standard deviation over trials. Note that we only try the interval size equal to 4 in this experiment.}
\label{Tab:tab1}
\end{table*}
\section{Related work}
Our active multi-step TD algorithm incorporates two key ingredients: active learning with objective which consists of TD errors and entropy maximization to enable stability and exploration, coarse-to-fine time chunking, and multi-step TD learning with an actor-critic architecture. In this section, we review previous works that draw on some of these topics.

The basic idea of actor-critic algorithms \cite{Witten77,Konda00} is that it learns the policy and value function simultaneously, where the actor takes action according to the current policy while the critic (or value function) estimates whether this action will improve the total returns or not. Further, the advantage actor-critic ($A^2C$) introduces the average value to each state, and leverages the difference between value function and the average to update the policy parameters. In other words, the good policy which leads to high value function will be enhanced and bad policy will be suppressed. Since the policy updates in each time step (or state), it can significantly reduce the variance compared to REINFORCE. %REINFORCE is effective, but it has high variance by delaying its model update until the end of the episode. The advantage of policy gradient algorithm is that it optimize the parameters related to the policy control, which in turn can maximize the total return. However, the major drawback is that they needs large number of training samples, which results in the inefficient usages of trajectories. In addition, policy based approaches have other weaknesses, such as high variance or bias. For example, REINFORCE algorithm leads to high variance because the gradient estimation requires the whole trajectory.

%To reduce the variance, Schulman et al proposed the generalized advantage value estimation \cite{Schulmanetal2016}, which considered the whole episode with an exponentially-weighted estimator of the advantage function that is analogous to $TD(\lambda)$  to substantially reduce the variance of policy gradient estimates at the cost of some bias

%The role of the critic is to evaluate the current policy prescribed by the actor. In principle, this evaluation can be done by any policy evaluation method commonly used, such as $TD(\lambda)$ [6], [18], LSTD [3], [18], [47] or residual gradients [25]. The critic approximates and updates the value function using samples. The value function is then used to update the actor policy parameters in the direction of performance improvement

How to balance the variance and bias is an interesting topic in the reinforcement learning. $TD(\lambda)$ \cite{Sutton1988} was first introduced by Sutton to update value function and learn policy control. The coefficient $\lambda \in [0, 1]$ trades off between bias and variance, and empirically shows that intermediate $\lambda$ performs best. For each state on the trajectory, $TD(\lambda)$ adjusts the parameter $\lambda$ to approximate target value function, where $\lambda=1$ is the Monte-Carlo return, i.e. sum of discounted future rewards. This is unbiased, but may have high variance because of the long stochastic sequence of rewards. For $\lambda=0$, the target value is updated using Bellman equation by sampled one-step lookahead. Unfortunately, this value may be biased because of the potential inaccurate estimation of value function. %This value has lower variance-i.e. single state transition- but is biased by the potential inaccuracy of the lookahead estimate of V. However, it , and it requires the user to manually tune a stepsize schedule for good performance.
Apparently, it is a challenge to achieve the good balance between bias and variance, because it requires the user to tune a stepsize manually. Late, it was extended with least squares TD in \cite{Bradtke96,Boyan02} to approximate the value function by summing the decayed estimators, in order to eliminate all stepsize parameters and improve data efficiency. Recently, Schulman et al. proposed a similar method, called the generalized advantage value estimation \cite{Schulmanetal2016}, which considered the whole episode with an exponentially-weighted estimator of the advantage function that is analogous to $TD(\lambda)$ to substantially reduce bias of policy gradient estimates. However, the actor-critic methods provides the low-variance baseline at the cost of the some bias and remain sample inefficient. Moreover, the actor-critic updates policy and value function at ``arbitrarily" every state, which slows down the whole training process especially for the deep network approximators. 

%A3C\cite{Mnih2016} to  learning that uses asynchronous gradient descent for optimization of deep neural network controllers
Another trend is to combine maximum entropy with policy gradient in reinforcement learning to balance exploration and exploitation. Maximum entropy inverse reinforcement learning \cite{Ziebart2008} was proposed to maximize the likelihood of the observed data with maximum entropy constraint. Relative Entropy Policy Search \cite{Peters2010} extends policy gradient under the relative entropy constraints, such as Kullback-Leibler divergence. Haarnoja et al. proposed soft Q-learning \cite{HaarnojaTAL17}, by incorporating maximum entropy into policy control, and showed it improved exploration and compositionality. Late, soft Q-learning is extend to soft actor critic \cite{HaarnojaZAL18}, an off-policy algorithm based on the maximum entropy, where the actor aims to maximize expected reward while also maximizing entropy.
% a more stable training procedure and improved performance by reducing approximation error variance in the target values 
% Actor-critic algorithms are typically derived starting from policy iteration, which alternates between policy evaluation—computing the value function for a policy—
%Because the continue-time variant has been developed to a reasonable level of maturity, this paper solely discusses algorithms in the discrete-time setting.  

Active learning has also been applied on reinforcement learning. For example, active inverse reinforcement learning \cite{macl09airl} was proposed that allows the agent to query the demonstrator for samples at specific states, instead of relying only on samples provided at ``arbitrary" states. Active Reinforcement Learning \cite{Epshteyn2008} is another method, which focuses on how policy is affected by changes in transition probabilities and rewards of individual actions, and then determine which states are worth exploring based solely on the prior MDP specification. Compared to previous active reinforcement learning \cite{Epshteyn2008}, our approach focuses on adaptively learning the step-size, while former method relies on the sensitivity of the optimal policy to the transitions and rewards. %It then focuses its exploration on the regions of space to which
Riad et al. proposed APRIL \cite{Akrour2012}, an active ranking mechanism, which combined with preference-based reinforcement learning in order to decrease the number of ranking queries to the expert needed to yield a satisfactory policy. Our approach takes a more similar strategy as \cite{macl09airl} to query critic-critic and select the most significant states and steps to update the actor-critic in the inner loop. In a sense, it is more related to the meta-learning \cite{FinnAL17,Shedivat18,Xu18}.

%Model-Agnostic Meta-Learning (MAML) \cite{FinnAL17} [Finn et al., 2017a, Finn and Levine, 2018, Finn et al., 2017b, Grant et al., 2018, Al-Shedivat et al., 2018] learns a good initialisation of the model that can adapt quickly to other tasks within a few gradient update steps.

For example, meta-policy gradient algorithm in \cite{Xu18d} adaptively learns a global exploration policy in deep deterministic policy gradient \cite{LillicrapHPHETS15} to speed up the learning process significantly. Meta-gradient reinforcement learning in \cite{Xu18} takes gradients w.r.t. the meta-parameters (such as  discount factor or bootstrapping parameter) of a return function. Our meta-learning approach learns the step size to improve sample efficiency and stabilize and speed up the learning process. But stepsize is discrete variable, which is hard to optimize use the prior gradient-based algorithm. Moreover, the step size should be adaptively changed to reflect the importance of states in the horizon. In this paper, we leverage active learning to query actor-critic and adaptively select the states and steps, which in turn are feedback to actor-critic to learn better policy and value function.  
%algorithm runs online, while interacting with a single environment, and successfully adapts the return to produce better performance.

\section{Conclusion}
In this paper, we propose a context-aware multi-step reinforcement learning method, which can actively select states and actions, and switch on/off future backups adaptively while updating model parameters based on the context change. Specifically, we introduce the intervals or chunks to actively select states and adaptive multi-step TD with a context-aware classifier to turn on/off future rewards while computing the target value. The adaptive multi-step TD can leverage the recent TD algorithms such as TD3, and moreover it generalized TD($\lambda$) to adaptively average different n-step returns with corresponding binary variables. Furthermore, our approach internalizes multi-step TD learning, and can leverage any advanced reinforcement learning models to improve performance. The experimental results demonstrate our approach is effective on off-policy tasks, especially in the continuous control setting. 

\bibliography{acpaper2019}
\bibliographystyle{aaai}

\end{document}